\documentclass[lettersize,journal]{IEEEtran}
\usepackage{amsmath,amsfonts}
\usepackage{algorithmic}
\usepackage{algorithm}
\usepackage{array}
\usepackage[caption=false,font=normalsize,labelfont=sf,textfont=sf]{subfig}
\usepackage{textcomp}
\usepackage{stfloats}
\usepackage{url}
\usepackage{verbatim}
\usepackage{graphicx}
\usepackage{cite}
\usepackage[colorlinks,linkcolor=blue]{hyperref}

\usepackage{amssymb}
\usepackage{booktabs}
\usepackage{tabularx}
\usepackage{colortbl}  
\usepackage{xcolor}
\usepackage{array}   
\definecolor{lightgray}{gray}{0.95}
\definecolor{color3}{gray}{0.95}
\definecolor{rouse}{rgb}{0.981,0.961,0.941}

\begin{document}

\title{FUN: A Focal U-Net Combining Reconstruction and Object Detection for Snapshot Spectral Imaging}

\author{Dahua Gao, Yubo Dong $^\star$, Anqi Li, Zhenyuan Lin, Ang Gao, Danhua Liu,  and Guangming Shi
\thanks{$^\star$: Corresponding author: Yubo Dong.}
\thanks{Dahua Gao, Anqi Li, Zhenyuan Lin, Ang Gao, Danhua Liu, and Guangming Shi are with the School of Artificial Intelligence, Xidian University, Xi'an 710126, China (e-mail: \{dhgao, dhliu, gmshi\}@xidian.edu.cn;  \{anqili, linzhenyuan, anggao\}@stu.xidian.edu.cn).}
\thanks{Yubo Dong is with Changzhi Medical College, the Engineering Research Centre for Intelligent Data Assisted Diagnosis and Treatment in Shanxi Province, and Uniwave Artificial Intelligence Technology Co., Ltd., Changzhi 046000, China (e-mail: ybdong98@163.com).}
}

\markboth{Journal of \LaTeX\ Class Files,~Vol.~14, No.~8, August~2021}%
{Shell \MakeLowercase{\textit{et al.}}: A Sample Article Using IEEEtran.cls for IEEE Journals}


\maketitle

\begin{abstract}

Conventional push-broom hyperspectral imaging suffers from slow acquisition speeds, precluding real-time object detection; in contrast, snapshot spectral imaging enables instantaneous hyperspectral images (HSIs) capture, making real-time object detection feasible — yet its potential is often compromised by time-consuming post-capture reconstruction. To address this issue, we propose the Focal U-shaped Network (FUN), a novel end-to-end framework that jointly performs HSI reconstruction and object detection via multi-task learning. FUN employs a shared U-shaped backbone, where reconstruction provides underlying spectral information while detection guides semantic-aware priors learning — facilitating mutually beneficial task interaction. Crucially, we introduce focal modulation, an efficient alternative to self-attention that modulates spatial and spectral features while reducing quadratic computational complexity, enabling a self-attention-free architecture for joint reconstruction and detection. Furthermore, we contribute a new HSI object detection dataset with 8712 annotated objects across 363 HSIs to facilitate evaluation of the proposed method. Experiments demonstrate that FUN achieves state-of-the-art performance on both tasks, using 40\% fewer parameters and 30\% less computation than recent alternatives — making it promising for future real-time edge deployment. The code and datasets are available: \url{https://github.com/ShawnDong98/FUN}.


\end{abstract}


\section{Introduction}

Hyperspectral imaging captures detailed scene information by dividing the spectrum into tens or hundreds of bands, which has been widely used in biomedical research \cite{med}, healthcare \cite{hyperskin}, remote sensing \cite{ssmsn, nhsiqa}, and industrial inspection \cite{industrial}. Particularly, in defense and surveillance, the ``spectral-spatial integration" property of HSIs combines spatial and spectral information, enabling the detection of camouflage targets by exploiting inherent spectral differences. However, the imaging speed of traditional techniques based on line-scanning or area-scanning methods is slow, thereby posing significant challenges for real-time monitoring and detection of dynamic targets.

Recently, the coded aperture snapshot spectral imaging (CASSI) system \cite{arce2013compressive, wang2015high} has gained attention because it enables snapshot spectral imaging in dynamic scenes. The CASSI system uses a coded aperture to encode and a disperser to multiplex a 3-D spectral scene into a 2-D measurement, allowing for real-time spectral data capture. However, conventional CASSI pipelines require reconstructing the full 3-D HSI from the 2-D measurement — a process that negates the inherent real-time advantage of snapshot imaging. To the best of our knowledge, no method has successfully performed object detection directly on CASSI measurements in an end-to-end, reconstruction-free manner — a critical gap for real-time defense and surveillance applications.

\begin{figure}[t]
	\begin{center}
		\begin{tabular}[t]{c} \hspace{-3.8mm} 
			\includegraphics[height=0.3\textwidth, width=0.45\textwidth]{figures/teaser_v20250922.pdf}
		\end{tabular}
	\end{center}
        \vspace{-3mm}
	\caption{\small Comparison of different object detection approaches on the CASSI system. (a) The first approach directly applies object detection to the measurement. (b) The second approach reconstructs the measurement into the HSI, then performs object detection on the reconstructed HSI. (c) Our method employs a U-shaped neural network to extract multi-scale features and leverages them to simultaneously perform  HSI reconstruction and object detection.}
	\label{fig:teaser}
	\vspace{-7mm}
\end{figure}

An intuitive yet underexplored approach is to perform object detection directly on the 2-D measurements. However, as shown in Fig. \ref{fig:teaser}(a), existing detectors — designed for natural RGB or reconstructed HSI — suffer a severe performance drop due to measurement degradations: the encoding process of the coded aperture leads to a loss of object information, while the 3-D to 2-D multiplexing distorts object shapes. To address this issue, we propose a Focal U-shaped Network (FUN) to simultaneously perform HSI reconstruction and object detection via multi-task learning. On the one hand, the objective of HSI reconstruction guides FUN to exploit underlying spectral information and restore object-related information hidden in degraded measurements, thereby benefiting the detection task. On the other hand, object detection provides heuristic object-level semantic priors and extracts more detailed and structural features, which in turn improve HSI reconstruction.

The FUN adopts a shared U-shaped backbone network to effectively extracts multi-scale features for both HSI reconstruction and object detection. U-Net \cite{unet} architecture has seen substantial success in a wide range of applications due to its capacity to capture intricate spatial and contextual information. Most end-to-end (E2E) neural networks \cite{l-net, tsa-net, mst, cst} for CASSI reconstruction also utilize a U-shaped architecture to ensure efficient feature extraction. Additionally, with the advancements in object detection, employing a backbone network to extract top-down features, in conjunction with a feature pyramid network (FPN) \cite{fpn} to handle bottom-up features, has become a standard paradigm. This methodology, too, aligns with a U-shaped network structure. Given the architectural similarities between HSI reconstruction and object detection tasks, we propose a unified U-shaped neural network as a shared backbone for both, thereby enabling more efficient reuse of neural network parameters and exploiting key features in both tasks.

To enhance feature representation while maintaining efficiency, we integrate focal modulation \cite{focalnet} —a lightweight alternative to self-attention—into the proposed FUN. Motivated by the distinct characteristics of HSIs, such as  local and non-local similarities \cite{dernn, dauhst} and low-rank properties \cite{dltr, lrsdn, sert, zhang1, zhang2, zhang3, zhang4, zhang5, tcsvt3}, we further tailor the focal modulation into two specialized modules: (1) Focal Spatial Modulation (FSM), which captures both local details and non-local structural cues through hierarchical depth-wise convolutions and gating; and (2) Low-Rank Spectral Modulation (LRSM), which leverages the intrinsic low-rank property of spectral signatures by maintaining a learnable global low-rank memory to retrieve and enhance spectral statistics.

To evaluate the effectiveness of the proposed FUN and further advance object detection in snapshot spectral imaging systems, we construct a new HSI object detection dataset, acquired using a GaiaField spectrometer. The dataset comprises $363$ HSIs, each with a spatial resolution of $696 \times 775$ pixels and $40$ spectral bands spanning $490 – 700$ nm. All images are densely annotated with bounding boxes across five object categories, resulting in a total of $8712$ annotated instances (averaging $24$ objects per image). Experimental results demonstrate that FUN achieves state-of-the-art (SOTA) performance on this dataset in both end-to-end HSI reconstruction and object detection, while also requiring fewer parameters and lower computational costs.

Overall, our contributions can be summarized as follows:
\begin{itemize}
    \item We propose a Focal U-shaped Network (FUN) to simultaneously perform HSI reconstruction and object detection via multi-task learning. To the best of our knowledge, this work is the first to explore the application of object detection on the CASSI system.
    \item We propose Focal Spatial Modulation (FSM) and Low-Rank Spectral Modulation (LRSM), which are integrated into FUN to enable efficient, self-attention-free modeling of spatial and spectral representations.
    \item FUN achieves SOTA performance in both end-to-end HSI reconstruction and object detection on the proposed dataset, which will be made public for future research.
\end{itemize}

\section{Related Work}

\subsection{Snapshot Spectral Imaging}

Snapshot spectral imaging systems \cite{arce2013compressive, dcd} enable real-time spectral imaging by capturing high-dimensional signals with low-dimensional sensors and solving the inverse problem to recover the desired signal. A representative system is the CASSI system \cite{arce2013compressive}, which employs a coded aperture and a disperser to modulate the 3-D spectral scene. Subsequently, the dispersed spectral signals are captured by a 2-D detector.

Based on CASSI, various reconstruction methods have been proposed, including model-based methods \cite{twist, gap-tv, desci}, end-to-end (E2E) neural networks \cite{l-net, tsa-net, mst, cst}, plug-and-play (PnP) methods \cite{lrsdn, lr2dp, yuan2020plug, tcsvt4}, and deep unfolding networks (DUNs) \cite{dgsmp, dauhst, rdluf, dernn, adrnn, tcsvt5, tcsvt6}. Model-based methods typically alternate between solving a data sub-problem and a prior sub-problem. When solving the prior sub-problem, model-based methods adopt hand-crafted priors to regularize the reconstruction process, including sparse-based priors \cite{gap-tv, ansr, chen2022exploring, chen2024fast}, non-local priors \cite{dauhst, dernn, nlm}, and low-rank priors \cite{dltr, lr2dp, lrsdn, chen2024thick}. However, the hand-crafted priors in model-based methods result in limited reconstruction quality, and the iterative optimization process leads to slow reconstruction speeds. PnP methods treat the prior sub-problem as a denoising problem and employ a pre-trained denoising neural network as a denoiser, leveraging the powerful denoising capabilities of deep neural networks (DNNs). Compared to plug-and-play methods, which use a shared DNN as the denoiser, DUNs not only assign an independent DNN for each iteration's denoiser but also introduce DNNs to solve the data sub-problem \cite{dauhst, rdluf, dernn}. Model-based methods, PnP methods, and DUNs have all demonstrated good performance. However, these optimization-based methods all require iterative processes. As the number of iterations increases, the reconstruction speed slows down, which undermines the merit of snapshot imaging. In contrast, E2E methods can achieve satisfactory reconstruction performance with a small computational cost.

\vspace{-2mm}
\subsection{Object Detection in Degraded Scenes}

Object detection is a fundamental task in computer vision, which has received widespread attention. With the development of object detection,  various methods have been proposed, including single-stage detectors like YOLO \cite{yolo} and RetinaNet \cite{retinanet}, two-stage detectors such as Faster R-CNNs \cite{faster_rcnn}, anchor-free detectors like FCOS \cite{fcos} and YOLO-X \cite{yolox}, and end-to-end detectors such as DETRs \cite{detr, deformable_detr}. These techniques generally perform well in clear visibility scenes. However, their effectiveness decreases significantly in degraded scenes. In the CASSI system, measurements suffer from degradation due to spatial modulation by the coded aperture and shape distortion caused by dispersion and integration. Such degradation hinders the application of modern detectors in the CASSI system.

Some works have explored object detection on degraded scenes, such as object detection in adverse weather \cite{ia_yolo, togethernet, liu2023image, dsnet, denet, fa_yolo}, low-light conditions \cite{ia_yolo, togethernet, liu2023image, denet}, and underwater scenarios \cite{liu2023image}. For instance, IA-YOLO \cite{ia_yolo} enhances images adaptively for improved detection using a differentiable image processing module, proving effective in foggy and low-light conditions. Similarly, DSNet \cite{dsnet} designs a dual-subnet network to improve object detection in foggy conditions by jointly learning visibility enhancement, object classification, and localization through a detection subnet (RetinaNet) and a restoration subnet, achieving superior performance on foggy datasets. Moreover, TogetherNet \cite{togethernet} integrates image restoration with object detection through dynamic enhancement learning to improve detection performance in adverse weather conditions. DENet \cite{denet} combines a detection-driven enhancement network, utilizing Laplacian pyramid decomposition and multiple enhancement modules, with a YOLO detector to enhance detection performance in both normal and adverse weather conditions. FA-YOLO \cite{fa_yolo} designs a hierarchical feature enhancement module, an adaptive receptive field enhancement module, and a deformable gated head to improve object detection performance in remote sensing under adverse weather conditions by adaptively enhancing feature-level information and enriching context.  However, object detection on the CASSI system has not yet been explored. Object detection and recognition are important applications in hyperspectral imaging, thus exploring object detection on the CASSI system is necessary.

\section{The Mathematical Model of CASSI}

\begin{figure}[]
    \centering
    \includegraphics[width=1.0\linewidth]{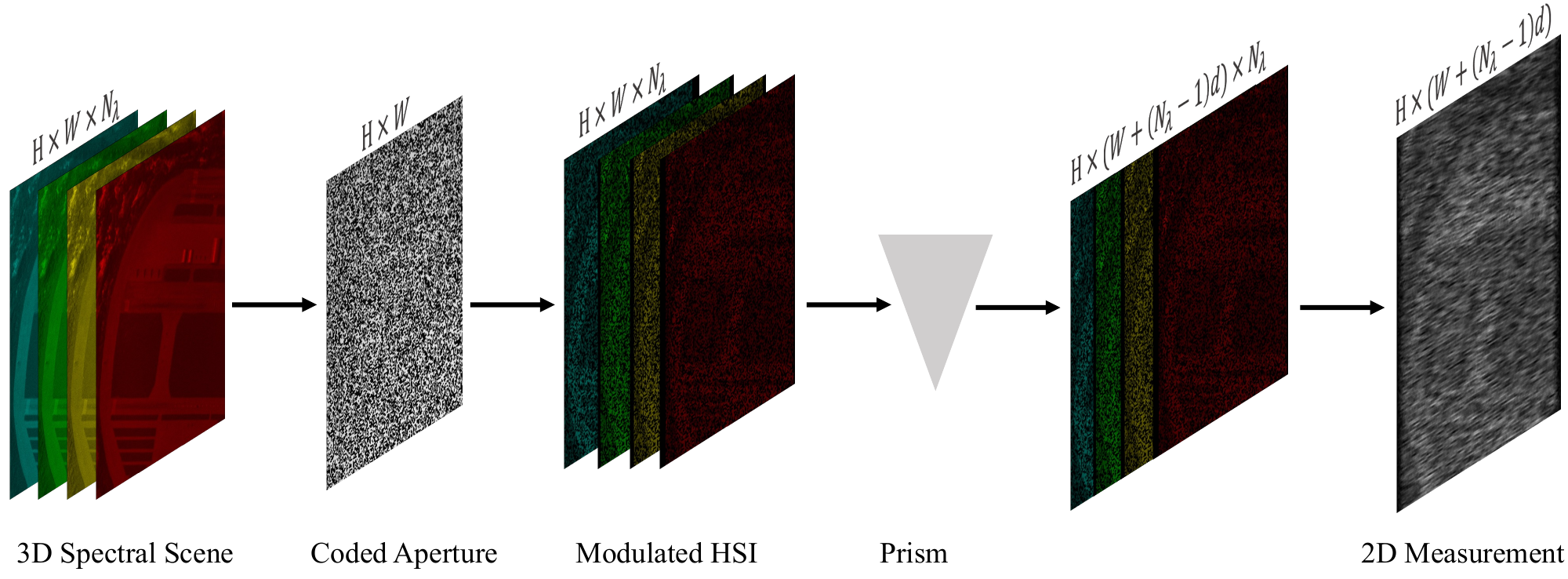}
    \caption{\small Diagram of the CASSI System. A 3-D spectral scene is first modulated by a coded aperture and a disperser (prism), then compressed into a measurement by a 2-D detector.} 
    \label{fig: CASSI}
\end{figure}

In the CASSI system \cite{arce2013compressive}, as shown in Fig. \ref{fig: CASSI}, a 3-D spectral scene is initially modulated by a coded aperture, which is a physical mask. Next, the modulated spectral information at different wavelengths is dispersed by a prism. Finally, the dispersed spectral components are captured by a 2-D imaging sensor and compressed into a single 2-D measurement.

Consider a 3-D spectral scene represented by an HSI signal tensor $X \in \mathbb{R}^{H \times W \times N_{\lambda}}$. Here, $H$ and $W$ denote the height and width respectively, while $N_{\lambda}$ represents the number of wavelengths. The coded aperture is depicted by a physical mask tensor $M \in \mathbb{R}^{H \times W}$. The process of modulating  the $n_{\lambda}$-th wavelength of the HSI can be expressed as:
\begin{equation}
    X_{n_\lambda}' = M \odot X_{n_\lambda},
    \label{eq: mask}
\end{equation}
where $\odot$ represents the Hadamard product (element-wise product). Then, a prism is used to spectrally disperse the spatially modulated HSI $X’$, which can be mathematically described as:
\begin{equation}
    X''(h, w+d_{n_\lambda}, n_\lambda) = X'(h, w, n_\lambda),
    \label{eq: disperse}
\end{equation}
where $X'' = \mathbb{R}^{H \times (W + d_{N_\lambda}) \times N_\lambda}$, with $d_{n_\lambda}$ representing the dispersed pixels corresponding to the $n_\lambda$-th wavelength. Finally, the dispersed HSI is compressed into a 2-D measurement by a 2-D detector. This process can described as: 
\begin{equation}
    Y = \sum_{n_\lambda = 1}^{N_\lambda} x_{n_\lambda}'' + N,
    \label{eq: sum}
\end{equation}
where $Y, N \in \mathbb{R}^{H \times (W + d_{N_\lambda})}$  respectively represent the 2-D measurement and the additive noise. The dispersion process increases the spatial dimensions of the signal, while the 2-D imaging process compresses the spectral dimension.

In snapshot spectral imaging, the typical goal is to solve the ill-posed problem of restoring the underlying 3-D HSI $x$ from the 2-D measurement $y$. However, there are few studies exploring the potential of utilizing the CASSI system for high-level computer vision tasks, such as object detection.

\section{The proposed Method}

\subsection{Focal U-Net}

\begin{figure}[]
    \centering
    \includegraphics[width=\linewidth]{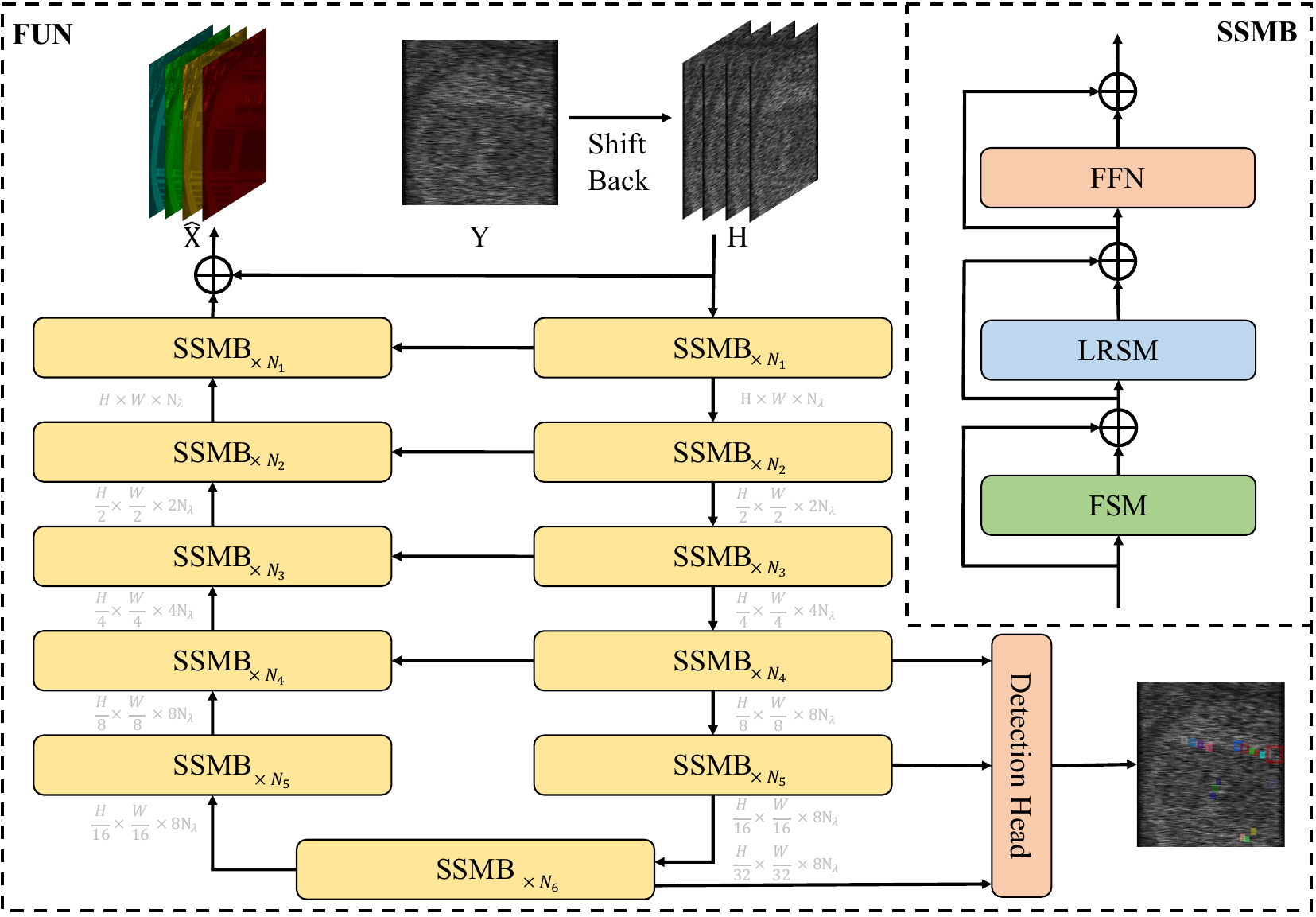}
    \caption{\small The overall architecture of FUN. FUN is a six-level U-shaped network with a lightweight detection head. Each level consists of $N_i$ SSMBs. The SSMB consists of an FSM, an LRSM, and an FFN, with residual connections and layer normalization between each pair of modules.} 
    \label{fig: FUN}
\end{figure}

The overall architecture of FUN is illustrated in Fig. \ref{fig: FUN}. FUN simultaneously performs a low-level HSI reconstruction task and a high-level object detection task. These tasks collaborate and promote each other. The proposed FUN is a 6-level U-shaped neural network with a lightweight detection head \cite{fcos}. Given a measurement $Y \in \mathbb{R}^{H \times (W + d_{N_\lambda})}$. Firstly, we reverse the dispersion process as described in Eq. (\ref{eq: disperse}), which shifts the measurement back to be used as the initial input:
\begin{equation}
    H(x, y, n_\lambda)= Y(x, y-d_{n_\lambda}).
    \label{eq: initial}
\end{equation}
The initial input $H \in \mathbb{R}^{H \times W \times N_\lambda}$  undergoes processing through six stages, each containing $N_i$ Spatial-Spectral Modulation Blocks (SSMBs). Between each pair of stages, downsampling and upsampling layers are applied.  The downsampling layer is a strided convolution  $2 \times 2$ layer that reduces the size of feature maps while doubling the number of channels. And the upsampling layer is a strided deconvolution  $2 \times 2$ layer that increases the size of feature maps and halves the number of channels. To reduce computational cost and memory usage, the number of channels is not increased beyond the fourth stage. After passing through all stages, the model generates a residual HSI denoted as $R$. The reconstructed HSI $\hat X$  is then obtained by adding the initial input $H$ to the residual $R$, expressed as $\hat X = H + R$.  Additionally, the output features of stages 4, 5, and 6 are processed by the lightweight detection head to generate the detection results.

The SSMB is the core component of FUN. SSMB consists of a Focal Spatial Modulation (FSM) module, a Low-Rank Spectral Modulation (LRSM) module, and a Feedforward Network (FFN), with LayerNorm and residual connections between each pair of modules. Both the FSM and LRSM are self-attention-free structures, which will be detailed in the following sections.

\subsection{Focal Modulation}

\begin{figure}[]
    \centering
    \includegraphics[width=\linewidth]{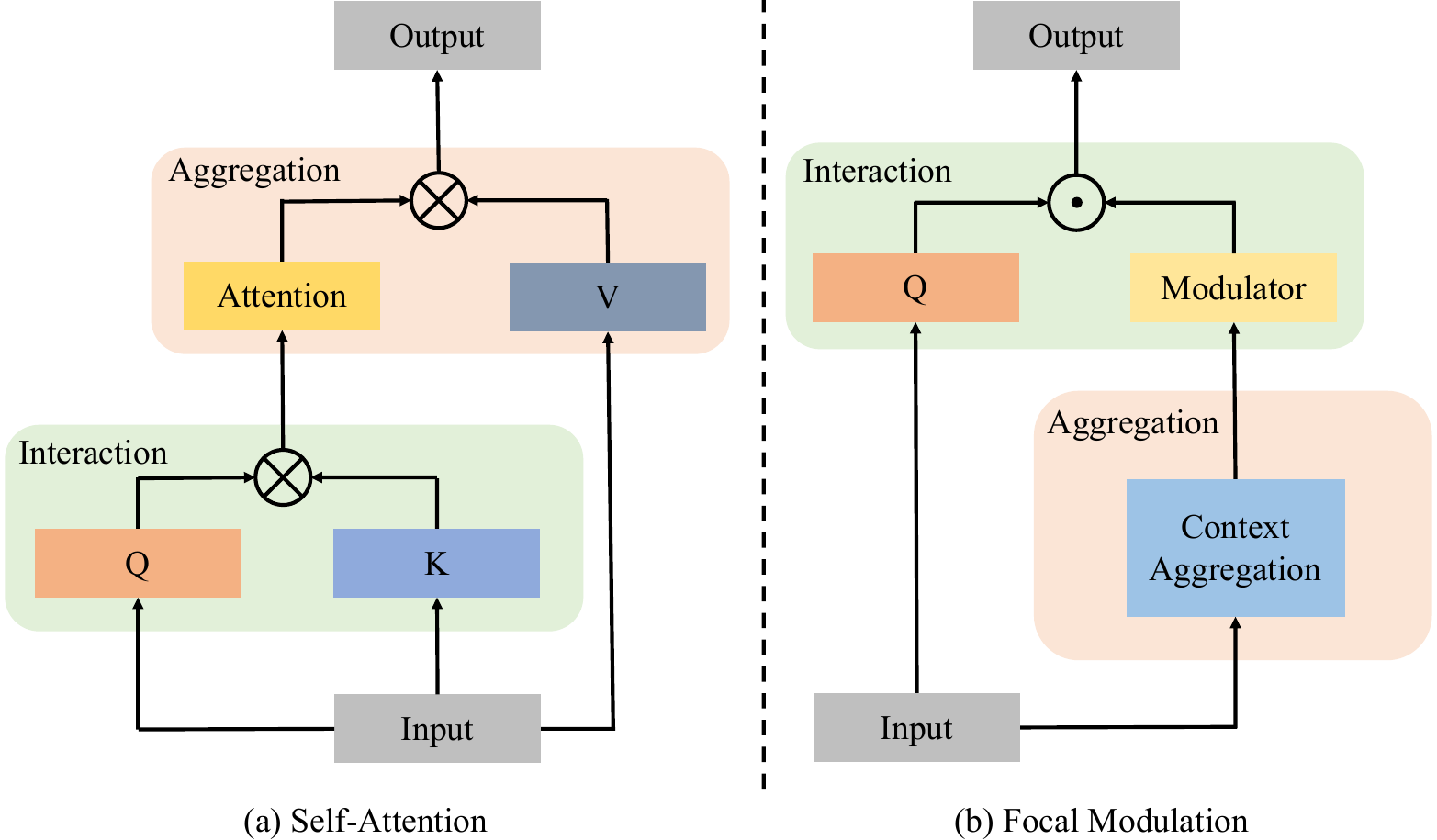}
    \caption{\small Comparison of Self-Attention and Focal Modulation. Self-Attention adopts a late aggregation approach, while Focal Modulation employs an early aggregation approach in representation learning.} 
    \label{fig: SAvsFM}
\end{figure}

A general representation learning process for a visual feature map $X \in \mathbb{R}^{H \times W \times C}$ involves generating a feature representation $y_i \in \mathbb{R}^C$ for each visual token (query) $x_i \in \mathbb{R}^C$, which is achieved by interacting with the surrounding tokens in $X$ (e.g., calculating the attention score) and then aggregating these contexts (e.g., the dot product between the value and the attention score).

\subsubsection{Self-Attention} As illustrated in Fig. \ref{fig: SAvsFM} (a), in self-attention, a late aggregation approach is employed, which can be expressed as:
\begin{equation}
    y_i = \text{aggrregation}(v(X),  \text{interaction}(q(x_i), k(X))),
    \label{eq: MSA}
\end{equation}
where the interaction first calculates the attention score between the query and key, and then the attention score is aggregated over the visual feature map $X$.

\subsubsection{Focal Modulation} As illustrated in Fig. \ref{fig: SAvsFM} (b), in focal modulation, the modulated representation $y_i$ is generated in an early aggregation approach, defined as:
\begin{equation}
    y_i = \text{interaction}(q(x_i), \text{aggregation}(i, X)),  
    \label{eq: FM}
\end{equation}
In this case, the contextual features are initially aggregated at each position $i$. Subsequently, the aggregated feature interacts with the query to generate the representation $y_i$.

Compared to Self-Attention, Focal Modulation has the following merits:
\begin{itemize}
    \item Focal Modulation simplifies aggregation by employing a shared operator, such as convolutions, which enhances computational efficiency. In contrast, Self-Attention requires intensive computation, as it necessitates summing non-shared attention scores for each query.
    \item The interaction in Focal Modulation is realized through an efficient operator, specifically an element-wise product between a token and its context. In contrast, Self-Attention computes attention scores between tokens, resulting in quadratic computational complexity.
\end{itemize}

Based on Eq. (\ref{eq: FM}), the Focal Modulation can be defined as:
\begin{equation}
    y_i = q(x_i) \odot m(i, X)
    \label{eq: FM1}
\end{equation}
where $\odot$ denotes element-wise multiplication and $q(\cdot)$ renpresents the query projection function. The function $m(\cdot)$ serves as the context aggregation, generating an output referred to as the modulator. 

\subsection{Focal Spatial Modulation}

\begin{figure}[]
    \centering
    \includegraphics[width=\linewidth]{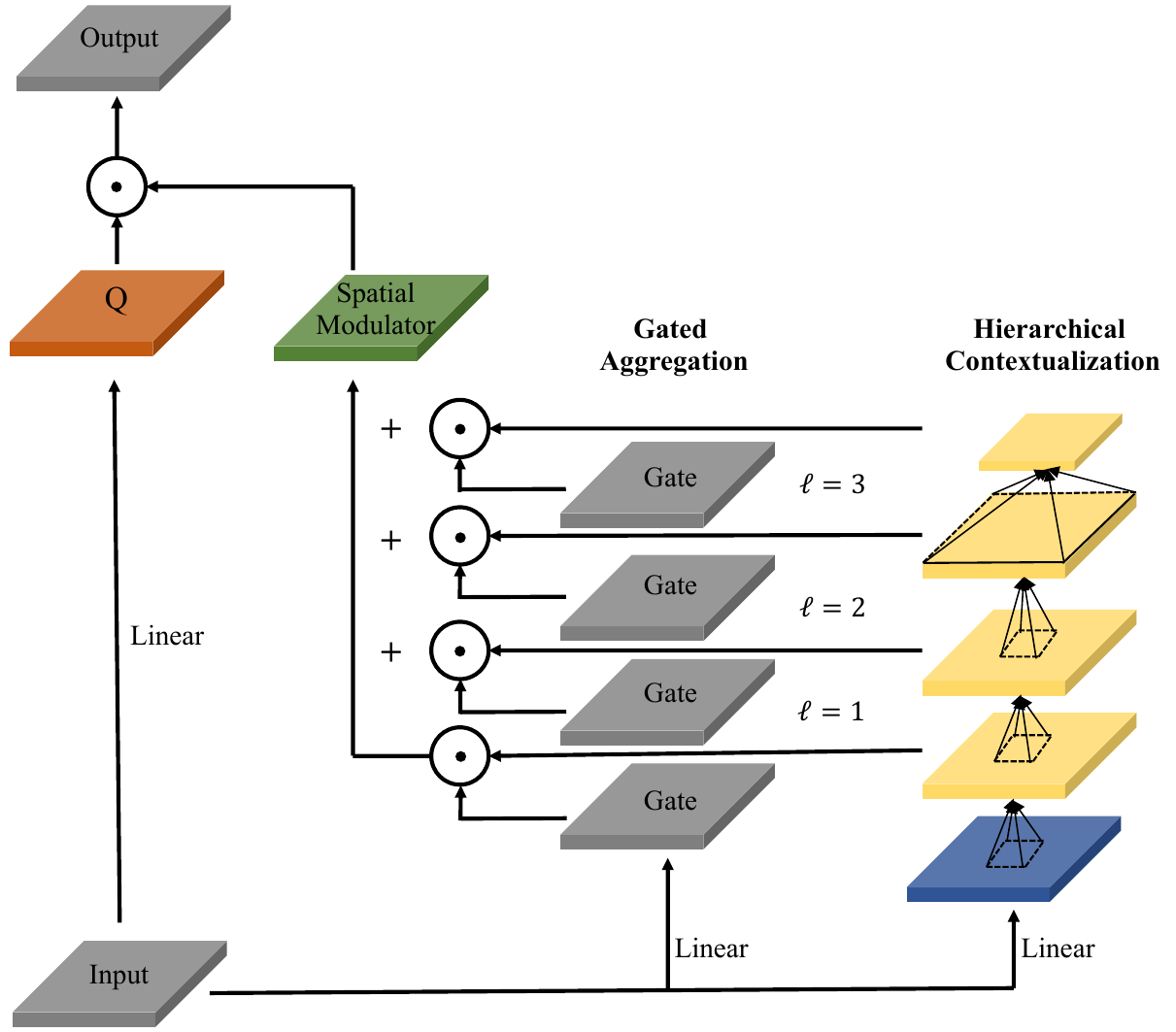}
    \caption{\small Diagram of the Focal Spatial Modulation (FSM). The FSM consists of two key components: Hierarchical Contextualization and Gated Aggregation. Hierarchical Contextualization extracts contextual information at various levels of detail, ranging from local to global. Gated Aggregation then integrates these extracted contextual features into the modulator for further processing.} 
    \label{fig: FSM}
\end{figure}

Both short- and long-range contexts play essential roles in visual modeling. However, a single aggregation with a large receptive field not only compromises local, fine-grained details, which are especially important for both reconstruction and object detection, but also leads to intensive computation and high memory consumption.

The Focal Spatial Modulation (FSM) employs hierarchical context aggregation across multiple scales. As illustrated in Fig. \ref{fig: FSM}, the aggregation involves two primary components: Hierarchical Contextualization and Gated Aggregation. Hierarchical Contextualization extracts context at varying levels of detail, from local to global. Gated Aggregation then integrates these contextual features into the modulator.

\subsubsection{Hierarchical Contextualization}  First, the feature map $\mathbf{X} \in \mathbb{R}^{H \times W \times C}$ undergoes a transformation into another feature space by applying a linear layer,  yielding $\mathbf{Z}^0 = f_z(\mathbf{X}) \in \mathbb{R}^{H \times W \times C}$.  Subsequently,  $L$ depth-wise convolutions are adopted to generate a hierarchical representation of contexts. The output $\mathbf{Z}^\ell$ is computed as follows:
\begin{equation}
    \mathbf{Z}^\ell = f_a^\ell(\mathbf{Z}^{\ell-1}) \triangleq \text{GeLU}(\text{DWConv}(\mathbf{Z}^{\ell-1})),
    \label{eq: Hierarchical Contextualization}
\end{equation}
where $\ell \in {1, \ldots, L}$ and $f_a^\ell$ denotes the contextualization operation at layer $\ell$. The contextualization operation is realized through a depth-wise convolution (DWConv) with a kernel size of $k^\ell$, and subsequently a GeLU activation is applied.

The effective receptive field is  $r^\ell = 1 + \sum_{i=1}^\ell (k^i - 1)$ at level $\ell$, which is significantly larger than the kernel size $k^\ell$. Then, a global average pooling is applied to the $L$-th level feature map to capture the global context of the input, which yields $\mathbf{Z}^{L+1} = \text{Avg-Pool}(\mathbf{Z}^L)$. As a result, we obtain $(L + 1)$ feature maps ${\mathbf{Z}^\ell}_{\ell=1}^{L+1}$, collectively capturing both short- and long-range contexts at varying scales.

\begin{figure}[]
    \centering
    \includegraphics[width=0.82\linewidth]{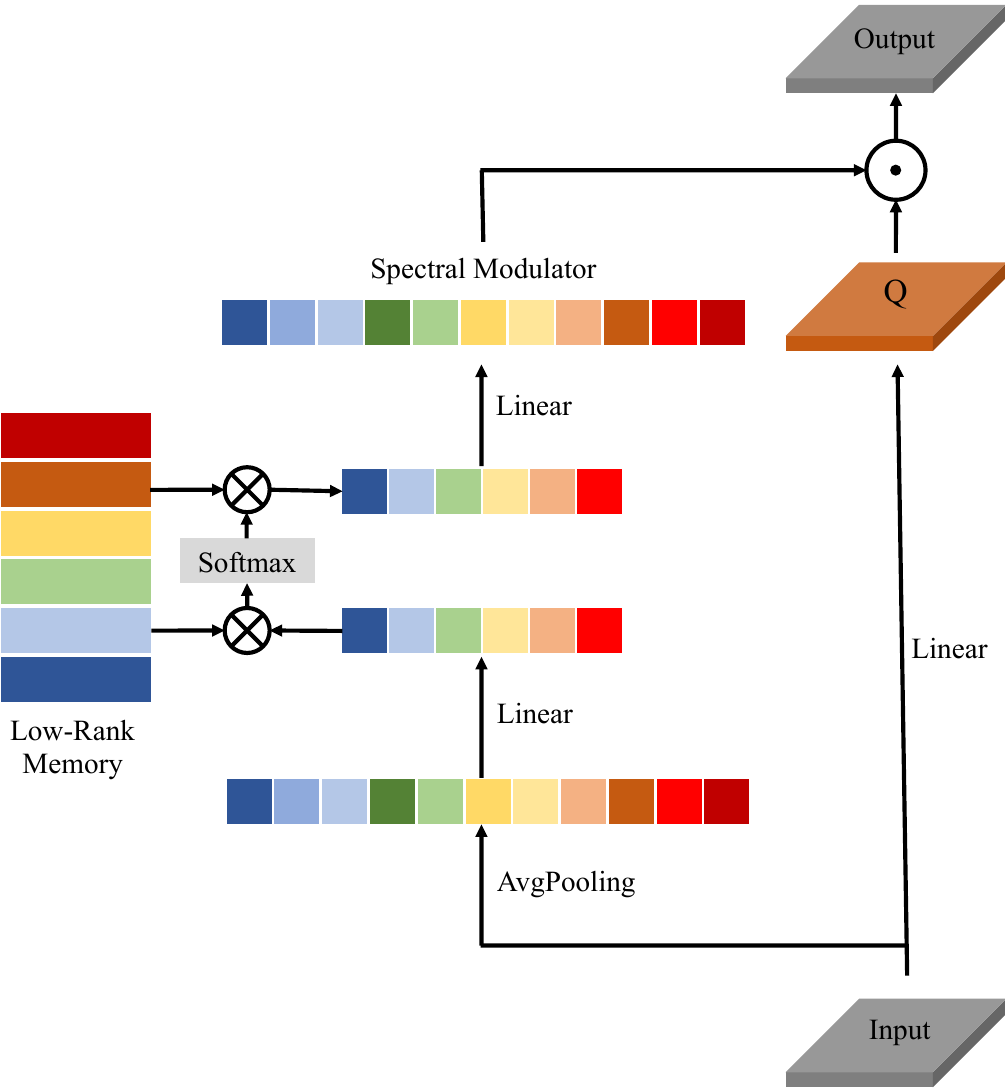}
    \caption{\small Diagram of the Low-Rank Spectral Modulation (LRSM). The LRSM projects the latent representation into a subspace to obtain a low-rank spectral vector and aggregates global spectral information from a low-rank memory to obtain an enhanced low-rank spectral vector.} 
    \label{fig: LRSM}
\end{figure}

\subsubsection{Gated Aggregation} Gated aggregation condenses $(L + 1)$ feature maps generated by the hierarchical contextualization into a modulator. The representation learning relies on local, fine-grained features to represent queries related to prominent visual objects, while global coarse-grained features may be used for background scenes. Therefore, a gated mechanism is employed to control the aggregation from different levels. Specifically, the gated mechanism uses a linear layer to compute gating weights $G = f_g(X) \in \mathbb{R}^{H \times W \times (L+1)}$. Subsequently, element-wise multiplication is applied to perform a weighted sum, obtaining a single feature map $Z^{out} \in \mathbb{R}^{H \times W \times C}$ :
\begin{equation}
    Z^{out} = \sum_{\ell=1}^{L+1} G^\ell \odot Z^\ell,
\end{equation}
where $G^\ell \in \mathbb{R}^{H \times W \times 1}$ represents the weight corresponding to level $\ell$. Finally, the modulator is obtained by applying a linear layer. 

\subsection{Low-rank Spectral Modulation}

The low-rank property of HSI has been widely utilized in various applications such as reconstruction \cite{lr2dp, lrsdn}, denoising \cite{sert}, and unmixing \cite{hsi_unmixing}, suggesting that a low-dimensional spectral subspace is beneficial for these tasks. However, projecting HSIs into an appropriate subspace can be challenging without robust regularization techniques like singular value decomposition (SVD). Rather than using orthogonal linear projection, we employ low-rank memory to capture the low-rank statistics of HSI. 

Low-Rank Spectral Modulation (LRSM) is illustrated in Fig. \ref{fig: LRSM},  which also involves two primary steps: Low-Rank Subspace Projection and Low-Rank Memory Aggregation. Low-Rank Subspace Projection projects the latent representation into a subspace to obtain a low-rank spectral vector. Low-Rank Memory Aggregation employs a low-rank memory to learn global spectral statistics in a data-driven manner and retrieves the most informative spectral statistics from the low-rank memory to form the enhanced low-rank spectral vector.

\subsubsection{Low-Rank Subspace Projection} The low-rank subspace projection applies a squeeze operation to aggregate the input feature map $\mathbf{X} \in \mathbb{R}^{H \times W \times C}$ into a spectral vector $Z^k \in \mathbb{R}^{1 \times 1 \times K}$. The squeeze operation includes a global average pooling and a linear projection. Firstly, an aggregated spectral vector $Z^c \in \mathbb{R}^{1 \times 1 \times C}$ is derived through a global average pooling operation. Then, the spectral vector $Z^c$ is projected into a subspace of rank $K$ to obtain $Z^k \in \mathbb{R}^{1 \times 1 \times K}$. The process of low-rank subspace projection is defined as:
\begin{equation}
    Z^k = \text{Linear}(\text{AveragePool}(X)).
    \label{eq: low-rank subspace projection}
\end{equation}

\begin{table*}[t]
	\renewcommand{\arraystretch}{1.5}
	\newcommand{\tabincell}[2]{\begin{tabular}{@{}#1@{}}#2\end{tabular}}
	\centering
        \caption{Comparisons between FUN and different methods are presented. The reported results include Params, GFlops, PSNR, SSIM, SAM, and mAP, with mAP calculated using IoU of 0.5.}
        \resizebox{\textwidth}{!}
	{
        \centering
        \begin{tabular}{c|c|c|c|c|c|c|c|c|c|c|c|c|c}
            \toprule[0.2em]
            \rowcolor{lightgray}
            ~~~~~Methods~~~~~
            & ~~~~~Params $\downarrow$~~~~~
            & ~~~~~GFlops $\downarrow$~~~~~
            & ~~~~~PSNR $\uparrow$~~~~~
            & ~~~~~SSIM $\uparrow$~~~~~
            & ~~~~~SAM $\downarrow$~~~~~
            & ~~~~mAP $\uparrow$~~~~
            \\
            \midrule
            RetinaNet \cite{retinanet}
            & 36.476M
            & 127.0
            & -
            & -
            & -
            & 44.4
            \\
            \midrule
            FCOS \cite{fcos}
            & 32.187M
            & 124.6
            & -
            & -
            & -
            & 45.7
            \\
            \midrule
            Faster RCNN \cite{faster_rcnn}
            & 41.432M
            & 133.5
            & -
            & -
            & -
            & 50.0
            \\
            \midrule
            Deformable DETR \cite{deformable_detr}
            & 41.432M
            & 133.5
            & -
            & -
            & -
            & 40.8
            \\
            \midrule
            L-Net \cite{l-net} + FCOS 
            & 64.926M
            & 530.3
            & 29.71
            & 0.661
            & 11.907
            & \underline{54.0}
            \\
            \midrule
            TSA-Net \cite{tsa-net} + FCOS 
            & 98.284M 
            & 1096.1
            & 27.61
            & 0.691
            & 14.100
            & 46.6
            \\
            \midrule
            MST \cite{mst} + FCOS 
            & 33.852M
            & 290.9
            & \underline{33.44}
            & \underline{0.776}
            & \underline{7.694}
            & 52.8
            \\
            \midrule
            CST \cite{cst} + FCOS
            & 34.619M
            & 385.1
            & 33.28
            & 0.772
            & 7.695
            & 49.5
            \\
            \midrule
            IA-YOLO \cite{ia_yolo}
            & 61.664M
            & \underline{112.8}
            & -
            & -
            & -
            & 30.67
            \\
            \midrule
            TogetherNet \cite{togethernet}
            & \bf{16.17M}
            & \bf{104.0}
            & 27.23
            & 0.719
            & 11.735
            & 34.73
            \\
            \midrule
            FUN
            & \underline{21.583M}
            & 270.2
            & \bf{33.66}
            & \bf{0.777}
            & \bf{7.473}
            & \bf{68.97}
            \\
            \bottomrule[0.2em]
        \end{tabular}}
	\label{Tab: quantitative results}
\end{table*}

\subsubsection{Low-Rank Memory Aggregation} The low-rank memory aggregation comprises a global memory bank $M \in \mathbb{R}^{K \times B}$, which are learnable parameters as part of the network. The low-rank memory aggregation identifies the most relevant low-rank spectral vectors from the spectral vector $Z^k$ and the low-rank memory bank $M$.  The coefficients $I \in \mathbb{R}^{1 \times B}$ corresponding between $Z^k$ and the stored low-rank memory $M$ are determined by:
\begin{equation}
    I = \text{Softmax}(Z^k M). 
    \label{eq: softmax}
\end{equation}
Using the coefficient matrix $I$, the local-rank memory retrieves the most informative spectral vectors to form the low-rank vector $Z^l \in \mathbb{R}^{1 \times 1 \times K}$:
\begin{equation}
    Z^l = IM.
    \label{eq: revert}
\end{equation}
Then, another linear layer projects $Z^l$ from the low-rank subspace to the original space:
\begin{equation}
    Z^{out} = \text{Linear}(Z^l).
    \label{eq: linear}
\end{equation}
Finally, a sigmoid is applied on the $Z^{out}$ to obtain the spectral modulator.

\subsection{Loss Function}

The total loss function includes a detection loss and a reconstruction loss, which can be formulated as:
\begin{equation}
    L_{total} =  L_{det} + \lambda L_{recon},
    \label{eq: loss}
\end{equation}
where $L_{det}$ refers to the detection loss and $L_{recon}$ refers to the reconstruction loss, and $\lambda$ is a balance parameter. We adopt the Charbonnier loss as the reconstruction loss, and the detection loss can be expressed by:
\begin{equation}
    L_{det} = L_{reg} + L_{cls} + L_{ctr},
\end{equation}
where $L_{reg}$ is the IoU loss, which is used for object localization. $L_{cls}$ is the focal loss, used for object classification. $L_{ctr}$ denotes the centerness loss, indicating the confidence that an object exists at a point.

\section{Experiments}

\subsection{Datasets and Experimental Settings}

\subsubsection{Datasets} To evaluate the effectiveness of FUN, we constructed an object detection HSI dataset with a GaiaField spectrometer, including a variety of terrains and scenes. The dataset will be made publicly available. To simulate a complex and varied terrain environment, we constructed simulation scenes with various types of terrain at a 1:700 scale, including scenarios such as deserts, airports, hills, ports, and oceans.  The dataset comprises 363 HSIs, annotated with five classes of objects: airplanes, ships, houses, stores, and trucks. Each HSI has a spatial resolution of $696 \times 775$.  The spectral range of the data spans from 490 nm to 700 nm, resulting in each HSI containing 40 channels. In total, the dataset includes 8712 objects, with an average of 24 objects per HSI. For simulating the CASSI measurement, a real physical mask is adopted and the dispersion shift step is set to 1.

\subsubsection{Experimental Settings} Experiments are implemented using PyTorch 1.12.1 on an RTX 4090 GPU. We employ the AdamW optimizer with hyperparameters $\beta_1 = 0.9$ and $\beta_2 = 0.999$, and set the weight decay to 0.0001. The training process spans a total of 90,000 steps, utilizing a multi-step scheduler with linear warm-up. The learning rate is set to 0.0001 and is reduced to 10\% of the original value at steps 60,000 and 80,000, respectively.

\begin{figure*}[]
    \centering
    \includegraphics[width=\linewidth]{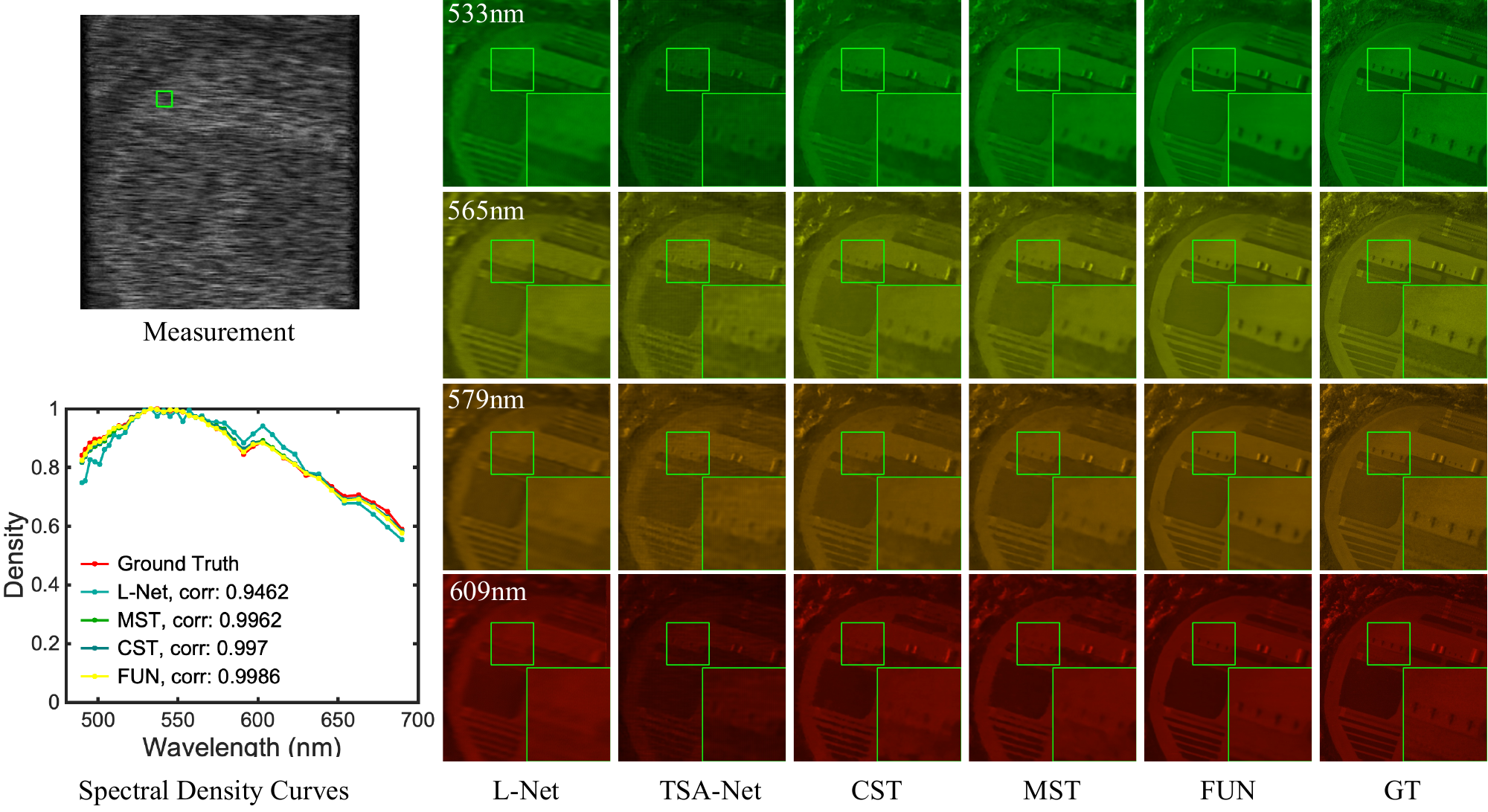}
    \caption{\small Comparisons of reconstructed HSIs across various wavelengths (533 nm, 565 nm, 579 nm, and 609 nm). The bottom-left shows the spectral curves corresponding to the green boxes in the measurement.  The right side displays the reconstructed HSIs using different methods, with the enlarged patches corresponding to the green boxes shown in the bottom-right corner of the reconstructed HSIs. Zoom in for a better view.} 
    \label{fig: ReconstructionResults}
\end{figure*}

\subsection{Quantitative Results}

\subsubsection{HSI Reconstruction} To assess the effectiveness of the proposed method for HSI reconstruction, we conducted a comparison with four end-to-end neural network approaches (L-Net \cite{l-net}, TSA-Net \cite{tsa-net}, MST \cite{mst}, and CST \cite{cst}). End-to-end neural networks prevent high computational complexity from the iterative processes in model-based methods, PnP techniques, and DUNs. The high computational complexity undermines the snapshot imaging of the CASSI system. The comparison results are shown in Tab. \ref{Tab: quantitative results}. With 21.583M parameters and 270.2 GFLOPs, FUN achieves the highest PSNR of 33.66 dB and the highest SSIM of 0.777 among the compared methods, indicating superior performance in terms of image quality and structural similarity. Additionally, FUN achieves the lowest SAM of 7.473, demonstrating its effectiveness in reconstructing spectra. Compared to TSA-Net, L-Net, MST, and CST, FUN improves the PSNR by 6.05 dB, 3.85 dB, 0.48 dB, and 0.22 dB, respectively, with fewer parameters and computational cost. These results highlight the efficiency of FUN in achieving high performance with relatively lower computational resources.

\subsubsection{HSI Object Detection} To assess the effectiveness of FUN in HSI object detection, we conducted a comparative analysis with ten different approaches,  including four object detection methods (a single-stage detector RetinaNet \cite{retinanet}, a two-stage detector Faster R-CNN \cite{faster_rcnn}, an anchor-free detector FCOS \cite{fcos}, and a Transformer-based detector Deformable DETR \cite{deformable_detr}), four ``reconstruction before detection'' methods (L-Net \cite{l-net}, TSA-Net \cite{tsa-net}, MST \cite{mst}, and CST \cite{cst} as the reconstructor, and FCOS \cite{fcos} as the detector), and two approaches specifically designed for degraded scene object detection (IA-YOLO \cite{ia_yolo} and TogetherNet \cite{togethernet}).  The comparison results are illustrated in Tab. \ref{Tab: quantitative results}. Compared to methods that directly apply object detection on measurements (RetinaNet \cite{retinanet}, FCOS \cite{fcos}, Faster R-CNN \cite{faster_rcnn}, and Deformable DETR \cite{deformable_detr}), FUN achieves higher detection performance on the mAP metric by 24.5, 23.2, 18.9, and 28.1, respectively, while only utilizing 59.17\%, 67.06\%, 52.20\%, and 53.74\% of parameters. Moreover, compared to the methods that combine reconstruction and detection (LNet \cite{l-net}, TSA-Net \cite{tsa-net}, MST \cite{mst}, CST \cite{cst} combined with FCOS \cite{fcos}), FUN achieves superior detection accuracy with fewer parameters and lower computational costs. Specifically, FUN improves the mAP metric by 14.9, 22.3, 16.1, 18.9, and 28.1 respectively, while it requires 33.57\%, 21.96\%, 64.76\%, 62.34\% fewer parameters and 50.95\%, 24.65\%, 92.88\%, and 70.16\% lower computational complexity.  Compared to methods designed for degraded scene object detection (IA-YOLO \cite{ia_yolo} and TogetherNet \cite{togethernet}), the mAP metric of FUN significantly outperforms IA-YOLO by 38.2 and TogetherNet by 34.2. In summary, FUN achieves the highest mAP score of 68.97, demonstrating a significant advantage in detection accuracy with fewer parameters and lower computational costs, making it a highly efficient and effective method compared to other approaches.

\begin{figure*}[]
    \centering
    \includegraphics[width=\linewidth]{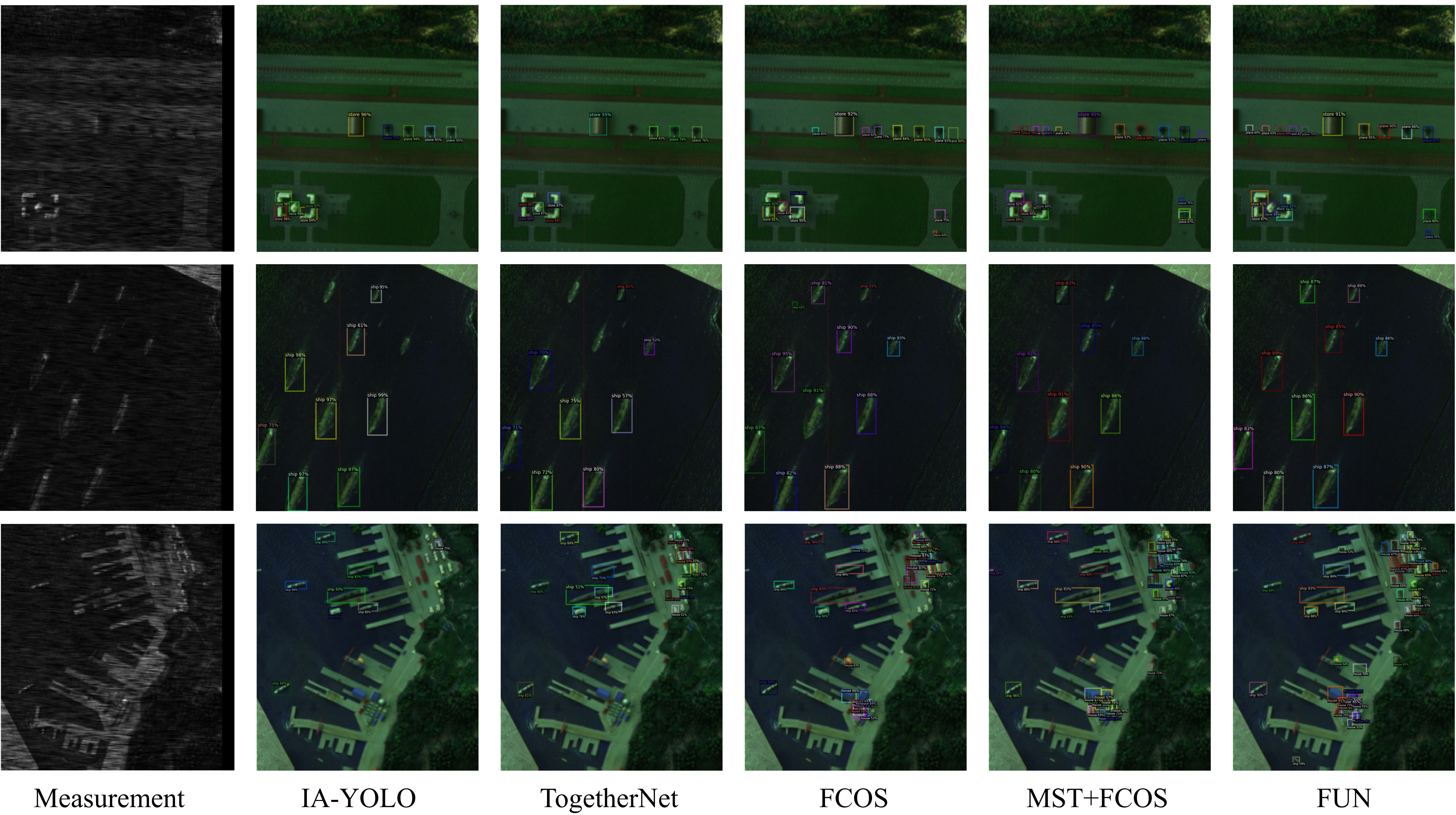}
    \caption{\small Detection results by various methods on three different scenes. The proposed FUN not only detects more objects with higher confidence, but also disentangles and identifies densely packed objects, maintaining high accuracy and minimizing overlap among bounding boxes.
    } 
    \label{fig: DetectionResults}
\end{figure*}

\subsection{Qualitative Results}

\subsubsection{HSI Reconstruction} For qualitative comparisons, as shown in Fig. \ref{fig: ReconstructionResults}, we present reconstruction results of FUN in comparison to L-Net \cite{l-net}, TSA-Net \cite{tsa-net}, CST \cite{cst}, and MST \cite{mst} across various wavelengths (533 nm, 565 nm, 579 nm, and 609 nm). As can be seen from the reconstructed HSIs (right), FUN consistently produces images that are closer to the ground truth (GT), exhibiting enhanced clarity and detail preservation. Specifically, the fine structures and textural details within the green boxes are more accurately reconstructed in our results, showcasing the robustness and precision of our approach. Benefiting from the object-level information brought by object detection, FUN clearly reconstructs all the airplanes. In contrast, the compared methods either show significant blurring and loss of detail or produce overly smooth results, leading to the loss of airplanes. The spectral density curves corresponding to the green box in the measurement are shown in the bottom-left corner. The highest correlation between our curve and the ground truth clearly demonstrates the effectiveness of FUN in restoring spectral consistency. These results highlight the effectiveness of FUN in achieving high-quality HSI reconstruction, thereby offering an improvement over existing techniques.

\subsubsection{HSI Object Detection} Fig. \ref{fig: DetectionResults} shows the qualitative comparisons between IA-YOLO \cite{ia_yolo}, TogetherNet \cite{togethernet}, FCOS \cite{fcos}, MST \cite{mst} + FCOS,  and FUN. In the first scene, which focuses on detecting objects in an airport setting, FUN clearly outperforms others by accurately identifying and bounding multiple objects with high precision. The bounding boxes predicted by FUN are more consistent and exhibit fewer false positives compared to other methods. The second scene involves the detection of ships in a maritime environment. FUN not only detects a higher number of ships but also provides more precise confidence scores, demonstrating its robustness and reliability. Other methods either miss several ships or produce bounding boxes with lower confidence levels. In the third scene, a complex port scenario with numerous overlapping objects, FUN again demonstrates superior performance. It successfully disentangles and identifies densely packed objects, maintaining high accuracy and minimizing overlap among bounding boxes. In contrast, other methods struggle with the complexity, often leading to missed detections or significant overlaps. Overall, these results highlight the advantages of FUN in terms of detection accuracy, precision, and robustness across various challenging scenarios.

\subsection{Ablation Study}

\begin{table*}[!htbp]
    \centering
    \caption{\small Ablation Analysis of Different Components.}
    \begin{tabular}{c c c c c c c c c c c c}
            \toprule
            Method & Detection & Reconstruction  &  FSM & LRSM & Params  & GFlops & PSNR $\uparrow$ & SSIM $\uparrow$ & SAM $\downarrow$ & mAP $\uparrow$ \\
            \midrule
            (a) & \checkmark              & -         & -    &   -  & 14.797M & 176.3  & -  & -     & - & 32.71 \\
            (b) & -              & \checkmark         & -    &   -  & 14.797M & 176.3  & 27.08  & 0.669     & 13.543 & - \\
            (c) & \checkmark              & \checkmark         & -    &   -  & 14.797M & 176.3  & 29.80  & 0.733     & 9.360 & 50.51 \\
            (d) & \checkmark              & \checkmark         & \checkmark    &   -  & 19.832M & 253.9 & 32.53  & 0.770     & 7.748  & 60.95 \\
            (e) & \checkmark              & \checkmark         & -    &   \checkmark  & 16.547M & 192.6  & 31.29 & 0.754     & 8.215 & 57.43  \\
            (f) & \checkmark              & \checkmark         & \checkmark    &   \checkmark  & 21.583M & 270.2  & \bf{33.66} & \bf{0.777}     & \bf{7.473}  & \bf{68.97} \\
            \bottomrule
    \end{tabular}
    \label{tab: ablation study}
\end{table*}

The proposed FUN demonstrates superior performance compared to SOTA methods. To further evaluate its effectiveness, we conducted a series of ablation experiments to dissect the contributions of various components, including multi-task learning, the Focal Spatial Modulation (FSM) module, and the Low-Rank Spectral Modulation (LRSM) module. 


\subsubsection{Effectiveness of reconstruction for detection}

To validate the effectiveness of the reconstruction task for object detection, all proposed components were removed from FUN, leaving only the object detection task to establish a baseline model (as shown in Tab.~\ref{tab: ablation study} (a)). Upon introducing multi-task learning, the results presented in Tab.~\ref{tab: ablation study} (c) were obtained. These results indicate that incorporating the reconstruction task significantly improves detection performance, with the mAP increasing from 32.71 to 50.51. This improvement suggests that the reconstruction task effectively guides the network in leveraging latent spectral information from the measurements, thereby enhancing object detection accuracy. Additionally, we observed that the standalone object detection task struggled to converge without the reconstruction task.

\subsubsection{Effectiveness of detection for reconstruction}

To validate the effectiveness of the object detection task in enhancing the reconstruction task, we removed all proposed components from FUN and retained only the reconstruction task to establish another baseline model, as shown in Tab.~\ref{tab: ablation study} (b). By introducing multi-task learning, we obtained the results in Tab.~\ref{tab: ablation study} (c). These results demonstrate that the object detection task improves the performance of the reconstruction task. Specifically, the PSNR increases from 27.08 to 29.80, SSIM improves from 0.669 to 0.733, and SAM decreases from 13.543 to 9.360. This enhancement suggests that the object detection task provides heuristic object-level priors and detailed structural cues that benefit the reconstruction task, thereby improving its overall performance. Additionally, we observed that the standalone object detection task struggled to converge without the reconstruction task.

\subsubsection{Effectiveness of FSM and LRSM}

To validate the effectiveness of FSM and LRSM, we incrementally added each component to the baseline in Method (c). As shown in Tab.~\ref{tab: ablation study} (d) and (e), introducing the FSM leads to improvements in both reconstruction quality (e.g., PSNR increases from 29.80 to 32.53) and detection performance (mAP increases from 50.51 to 60.95). Incorporating the LRSM further enhances all evaluation metrics. Notably, combining both FSM and LRSM (Method (f)) yields the best overall performance, achieving the highest PSNR (33.66), SSIM (0.777), lowest SAM (7.473), and highest mAP (68.97), which confirms the complementary contributions of these two modules. Although the full model introduces a moderate increase in parameter count and computational complexity, the substantial performance gains justify this overhead.

\subsubsection{Choice of the loss balance parameter}

To investigate the influence of the loss balance parameter $\lambda$ on the model performance, we conducted an ablation study with $\lambda$  varying from 1 to 10. As shown in Tab. \ref{tab: ablation_lambda}, the model achieves the best overall performance when $\lambda$  is set to 5, yielding the highest PSNR (33.66) and SSIM (0.776), the lowest SAM (7.473), and the highest mAP (68.97). These results suggest that $\lambda=5$  offers an optimal trade-off between reconstruction quality and detection accuracy. Setting $\lambda$  either too low or too high leads to suboptimal performance, highlighting the importance of proper balance in multi-task learning.

\begin{table}[!htbp]
    \centering
    \caption{\small Ablation Study of the Balance Parameter $\lambda$.}
    \begin{tabular}{c c c c c c}
            \toprule
            & Balance parameter $\lambda$  & PSNR $\uparrow$ & SSIM $\uparrow$ & SAM $\downarrow$ & mAP $\uparrow$ \\
            \midrule
            (a) & 1  & 31.23 & 0.750  & 8.282 & 47.43 \\
            (b) & 3  & 32.59 & 0.766  & 7.784 & 59.57 \\
            (c) & 5  & \bf{33.66} & 0.\bf{776}  & \bf{7.473} & \bf{68.97}  \\
            (d) & 7  & 33.35 & 0.775  & 7.615 & 64.43  \\
            (d) & 10 & 33.04 & 0.771 & 7.688 & 63.01  \\
            \bottomrule
    \end{tabular}
    \label{tab: ablation_lambda}
\end{table}

\section{Conclusion}

In this paper, we proposed FUN to simultaneously perform HSI reconstruction and object detection via multi-task learning. The proposed FUN shares a U-shaped backbone to extract deep features at various scales, and performs different tasks by leveraging scale-specific features.  In this way, HSI reconstruction guides the FUN exploiting latent spectral information from the measurements, thereby enhancing object detection performance. Conversely, object detection provides crucial object-level semantic priors for the HSI reconstruction. Additionally, FUN introduces focal modulation to modulate spatial and spectral feature representations, resulting in a self-attention-free network. Specifically, we present the FSM to aggregate spatial features with hierarchical depth-wise convolutions and a gated mechanism, and the LRSM to aggregate spectral features from a data-driven low-rank memory, both of which modulate the query with the aggregated features to make it more representative. Experimental results validated the effectiveness of the proposed FUN.

\bibliographystyle{IEEEtran}
\bibliography{references}

\end{document}